\def\year{2020}\relax
\documentclass[letterpaper]{article} 
\usepackage{aaai20}  
\usepackage{times}  
\usepackage{helvet} 
\usepackage{courier}  
\usepackage[hyphens]{url}  
\usepackage{graphicx} 
\usepackage{graphicx}  
\usepackage{amsmath}
\usepackage{dsfont}
\usepackage{amssymb}
\usepackage{color}

\newcommand{\figref}[1]{Figure~\ref{#1}}
\newcommand{\tabref}[1]{Table~\ref{#1}}
\newcommand{\equref}[1]{Eq.~\ref{#1}}

\title{Hierarchical Reinforcement Learning as a Model of Human Task Interleaving}
\author{Christoph Gebhardt\textsuperscript{\rm 1}, Antti Oulasvirta\textsuperscript{\rm 2}, Otmar Hilliges\textsuperscript{\rm 1}\\
ETH Z\"urich\textsuperscript{\rm 1}, Aalto University\textsuperscript{\rm 2}}
 
\begin{document}

\maketitle

\begin{abstract}
How do people decide how long to continue in a task, when to switch, and to which other task?
Understanding the mechanisms that underpin task interleaving is a long-standing goal in the cognitive sciences.
Prior work suggests greedy heuristics and a policy maximizing the marginal rate of return.
However, it is unclear how such a strategy would allow for adaptation to everyday environments that offer multiple tasks with complex switch costs and delayed rewards. 
Here we develop a hierarchical model of supervisory control driven by reinforcement learning (RL).
The supervisory level learns to switch using task-specific approximate utility estimates, 
which are computed on the lower level. 
A hierarchically optimal value function decomposition can be learned from experience,
even in conditions with multiple tasks and arbitrary and uncertain reward and cost structures.
The model reproduces known empirical effects of task interleaving. 
It yields better predictions of individual-level data than a myopic baseline in a six-task problem ($N$=211).
The results support hierarchical RL as a plausible model of task interleaving.
\end{abstract}

\section{Introduction}

\begin{figure}[t]
\centering
\includegraphics[width=1.0 \columnwidth]{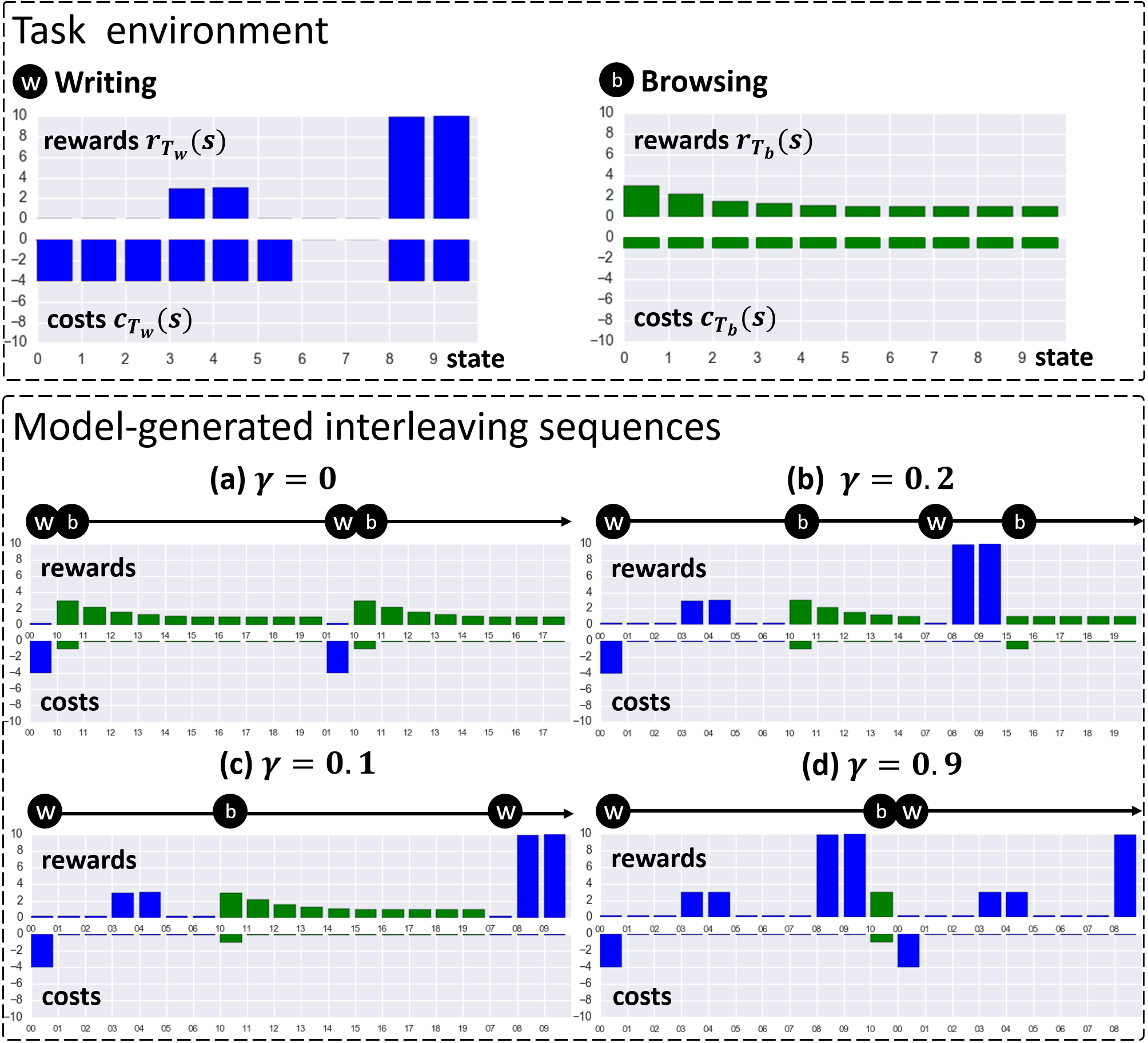}
\caption{Example of the task interleaving problem with two tasks: Given a limited time window and $N$ tasks with reward/cost structures, an agent has to decide what to focus on at any given time. Attending a task progresses its state and collects the associated rewards, while switching to another task incurs a cost. Shown are interleaving sequences (a-d) generated by a hierarchical reinforcement learner. Discount factors $\gamma$ specify the length of the RL reward horizon. }

\label{fig:toyproblem}
\end{figure}

How long will you keep reading this paper before you return to email? 
Knowing when to persist and when to do something else is a hallmark of cognitive functioning and is intensely studied in the cognitive sciences \cite{Altmann2002,Brumby2009,Duggan2013,Janssen2010,Jersild1927,monsell2003,Norman1986,Oberauer2011,Payne2007,Wickens2008}.
In the corresponding decision problem, 
\emph{the task interleaving problem}, 
an agent must decide how to share its resources among a set of tasks over some period of time.
We here investigate \emph{sequential} task interleaving, 
where only one demanding task is processed at a time.\footnote{For models of concurrent multitasking that involves simultaneous resource-sharing, see \cite{brumby2018,Oberauer2011,Salvucci2008}.}
The agent can focus on a task, thus advancing it and collecting its associated rewards.
It can also switch to another task, but this incurs a switch cost, 
the magnitude of which depends on the agent's current state \cite{Jersild1927,monsell2003}.
Consider the two-task interleaving problem shown in Figure \ref{fig:toyproblem}:
How would you interleave and how would a rational agent behave?
The general problem is non-trivial: 
our everyday contexts offer large numbers of tasks with complex and uncertain properties.  

It is well-known that interleaving behavior is adaptive. 
In particular, the timing of switches shows sensitivity to in-task rewards \cite{Horrey2006,janssen2015strategic,Iani2007,Wickens2008} 
and to resumption costs \cite{Altmann2002,Gutzwiller2019,Iqbal2008},
which are affected by skill-level \cite{janssen2015strategic} and memory recall demands \cite{Altmann2007,Oulasvirta2006}.
Task switches tend to be pushed to boundaries between tasks and subtasks, where switch costs are lower \cite{Altmann2002,janssen2012natural,McFarlane2002}.
Previous models have shed light on possible mechanisms underlying these effects: 
(i) According to a time-based switching heuristic, 
the least attended task receives resources, to balance resource-sharing among tasks \cite{Salvucci2008,Salvucci2009}, or in order to refresh it in memory \cite{Oberauer2011};
(ii) According to a foraging-based model, 
switching maximizes in-task reward \cite{Payne2007,Duggan2013}, 
which is tractable for diminishing-returns reward functions using the marginal value theorem;
(iii) According to a multi-attribute decision model,
task switches are determined based on task attractiveness, defined by importance, interest, and difficulty \cite{Wickens2015}. 

While these models have enhanced our understanding of interleaving,
we still have an incomplete picture of mechanisms underpinning \emph{adaptation}.
In order to fully explain human interleaving capabilities, 
we need to understand how interleaving adapts to multiple tasks and complex reward/cost structures,
including delayed rewards.
Examples with non-diminishing rewards are easy to construct:
in food preparation, the reward is collected only after cooking has finished.
In addition, we also need to explain people's ability to interleave tasks they have not experienced before.

Here we propose \emph{hierarchical reinforcement learning} (HRL) as a unified account of adaptive supervisory control in task interleaving.
While there is extensive work on HRL in machine learning, 
we develop it here specifically as a model of \emph{human} supervisory control that keeps track of on-going tasks and decides which to share resources to \cite{Norman1986,Wickens2008}.
We assume a two-level supervisory control system,
where both levels use RL to approximate utility based on experience.
RL in general is a plausible mechanism for utility approximation in conditions that are non-stationary, uncertain, and where gratifications are delayed \cite{Sutton1998}. 
In task interleaving, it models how people estimate the value of continuing in a task 
and can anticipate a high future reward even if the immediate reward is low.
\emph{Hierarchical} RL extends this by employing temporal abstractions that describe state transitions of variable durations.

Hierarchicality has cognitive appeal thanks to its computational tractability.
Selecting among higher-level actions reduces the number of decisions required to solve a problem \cite{Botvinick2012}.
We demonstrate significant decreases in computational demands when compared to an equal but flat agent. 
But HRL is also aligned with neuroscientific evidence, 
according to which the prefrontal cortex (PFC) is organized hierarchically for supervisory control \cite{Botvinick2012,Frank2011},
with dopaminergic signaling contributing to temporal-difference learning and PFC representing currently active subroutines.
As a consequence, HRL has been applied to explain brain activity during complex tasks \cite{Botvinick2009,Rasmussen2017,Balaguer2016}.
However, no related work considers hierarchically optimal problem decomposition of cognitive processes in task interleaving.
Hierarchical optimality is crucial in the case of task interleaving, since rewards of the alternative tasks influence the decision to continue the attended task.

This paper presents a novel computational implementation of HRL for task interleaving 
and assess it against a rich set of empirical findings.
The defining assumption of the model is a two-level hierarchical decomposition of the RL problem.
(i) On the lower -- or task type -- level, 
a state-action-value-function is kept for each \emph{task type} (e.g., writing, browsing) and updated with experience of each ongoing \emph{task instance} (e.g., writing task A, browsing task B, browsing task C).
(ii) On the higher -- or task instance -- level, 
a pointer is kept to each on-going task instance.
RL decides the next task based on value estimates provided from the lower level.
This type--instance distinction permits taking decisions without previously experiencing the particular task instance. 
By modeling task-type-level decisions with a \emph{semi-Markov Decision Process} (SMDP), we model how people decide to switch at \emph{decision points} rather than at a fixed sampling interval.
In addition, the HRL-model allows learning arbitrarily shaped reward- and cost functions.

We report evidence from simulations and empirical data. 
The model reproduces known patterns of adaptive interleaving
and predicts individual-level behavior measured in a challenging and realistic interleaving study with six tasks ($N$=211).
The HRL model was better or equal than a myopic baseline model, which does not consider long-term rewards.
HRL also showed more human-like patterns, such as sensitivity to subtask boundaries.
We conclude that human interleaving behavior appears better described by optimal planning under uncertainty than by a myopic strategy, 
and that hierarchical decomposition is a plausible cognitive solution to this planning problem. 
\section{Background}

\subsection{Markov and semi-Markov decision processes}
\label{sec:mdp}
The family of Markov decision processes (MDP) is a mathematical framework for decision-making in stochastic domains \cite{kaelbling1998}. An MDP is a four-tuple ($S$, $A$, $P$, $R$), where $S$ is a set of states, $A$ a set of actions, $P$ state transition probability for going from a state $s$ to state $s'$ after performing action $a$ (i.e., $P(s'|s,a)$),
and $R$ the reward for action $a$ in state $s$ (i.e. $R:S \times A \rightarrow \mathbb{R}$). 
The expected discounted reward for action $a$ in $s$ and then following policy $\pi$ is known as the Q value: $Q^{\pi}(s,a) = E_{s_t}[\sum_{t=0}^{\infty}\gamma^t R(s_t, a_t)]$ where $\gamma$ is a discount factor. 
Q values are related via the Bellman equation: $Q^\pi(s,a) = \sum_{s'} P(s'|s,a)[R(s',s,a) + \gamma Q^\pi(s',\pi(s'))]$.
The optimal policy can then be computed as $\pi^* = arg~max_a Q^\pi(s,a)$.
Classic MDPs assume a discrete step size. 
To model temporally extended actions, semi-Markov decision processes (SMDPs) are used. SMDPs represent snapshots of a system at decision points where the time between transitions can be of variable temporal length. 
An SMDP is a five-tuple ($S$, $A$, $P$, $R$, $F$), where $S$, $A$, $P$, $R$ describe an MDP and $F$ gives the probability of transition times for each state-action pair. 
Its Bellman equation is: $Q^\pi(s,a) = \sum_{s',t} F(t|s,a)P(s'|s,a)[R(s,a) + \gamma^t Q^\pi(s',\pi(s'))]$, 
where $t$ is the number of time units after the agent chooses action $a$ in state $s$ and $F(t|s,a)$ is the probability that the next decision epoch occurs within $t$ time units. 

\subsection{Hierarchical reinforcement learning}
\label{sec:hrl}
Hierarchical RL (HRL) is based on the observation that a variable can be irrelevant to the optimal decision in a state even if it affects the value of that state \cite{Dietterich1998}. 
The goal is to decompose a decision problem into subroutines, encapsulating the internal decisions such that they are independent of all external variables other than those passed as arguments to the subroutine. 
%
There are two types of optimality of policies learned by HRL algorithms. A policy which is optimal with respect to the non-decomposed problem is called \emph{hierarchically optimal} \cite{Andre2002,Ghavamzadeh2002}. 
A policy optimized within its subroutine, ignoring the calling context, is called \emph{recursively optimal} \cite{Dietterich1998}. 
\section{Computational Model of Task Interleaving}

\begin{figure}[b!]
\centering
\includegraphics[width=0.7 \columnwidth]{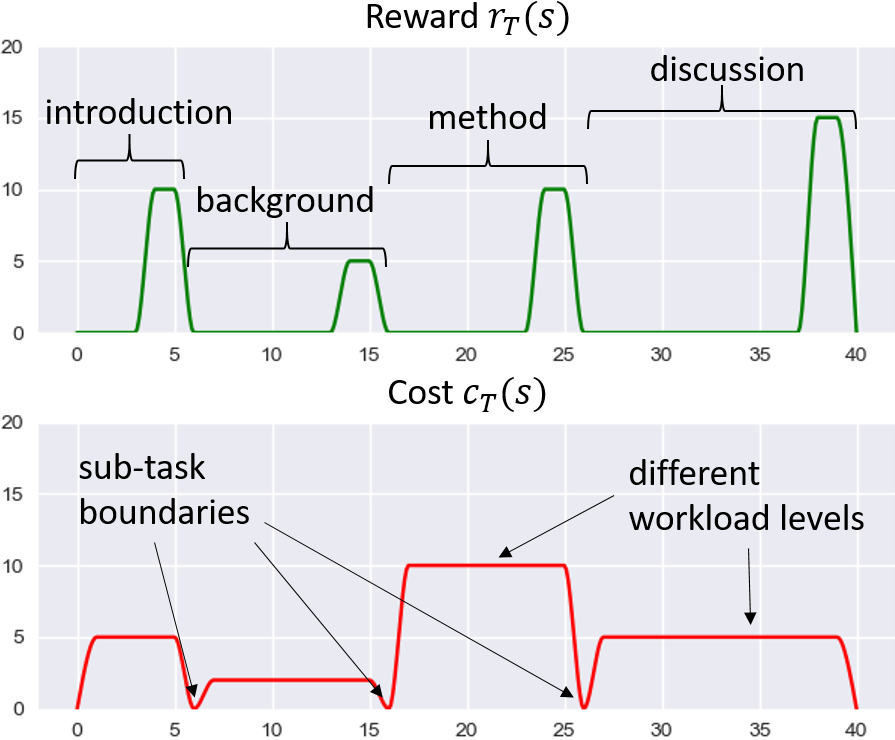}
\caption{An exemplary task model for paper writing: rewards $r_T$ and costs $c_T$ over discrete states $s$.}
\label{fig:task_model}
\end{figure}

\paragraph{Task model:}
We model tasks via the reward $r_T(s)$ and cost $c_T(s)$ functions defined over discrete states $s$ (see \figref{fig:task_model}). 
The reward represents subjective attractiveness of a state in a task \cite{Norman1986,Wickens2008}. 
The cost represents overheads caused by a switch \emph{to} a task \cite{Jersild1927,Oberauer2011,Oulasvirta2006}. 
A state is a discrete representation of progress within a task and the progress is specific to a task type. 
For instance, in our reading task model, 
progress is approximated by the position of the scroll bar in a text box. 
Reward and cost functions can be arbitrarily shaped.
This affords flexibility to model tasks with high interest \cite{Horrey2006,Iani2007}, tasks with substructures \cite{Bailey2006,Monk2004}, and complex resumption costs \cite{Rubinstein2001,Bailey2006}.

\begin{figure}[t!]
\centering
\includegraphics[width=0.95 \columnwidth]{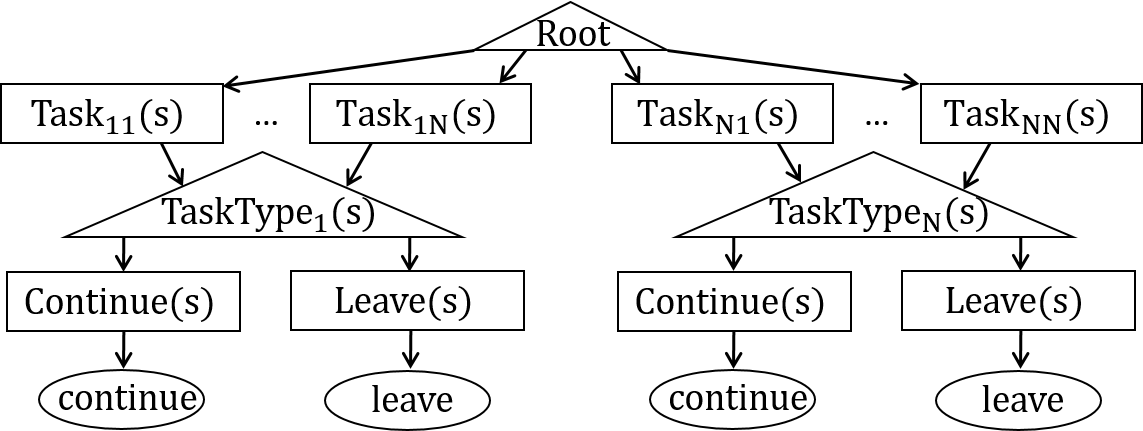}
\caption{A hierarchical decomposition of the task interleaving problem: subroutines are triangles, rectangles are composite actions and primitive actions are ovals. $Root$ chooses among all available task instances, e.g., $Task_{11}(s)$, which in turn call the subroutine of their respective type, e.g., $TaskType_1(s)$. A subroutine can either continue $Continue(s)$ or leave $Leave(s)$ a task.}
\label{fig:hrl}
\end{figure}

\paragraph{Hierarchical decomposition of task environments:}
We assume that interleaving requires generalization of an encountered task instance to its respective type. 
The environment of an agent can consist of multiple tasks, some of which share the same type (i.e., multiple task instances of type reading). 
\figref{fig:hrl} shows the hierarchical decomposition of the problem.
Rectangles represent composite actions that can be performed to achieve their parent's subroutine or call a primitive action (ovals). 
Each subroutine (triangle) is a separate SMDP.
The problem is decomposed by defining a subroutine for each task type: $TaskType_1(s)$ to $TaskType_N(s)$. 
A subroutine estimates the expected cumulative reward of pursuing a task from a starting state $s$ until the state it expectedly leaves the task. At a given state $s$, it can choose from the actions of either continuing $Continue(s)$ or leaving $Leaving(s)$ the task. 
These actions then call the respective action primitives: $continue$, $leave$. 
The higher level routine $Root$, selects among all available task instances, $Task_{11}(s)$ to $Task_{NN}(s)$, the one which returns the highest expected reward.
When a task instance is selected, it calls its respective task type subroutine passing its in-task state $s$ (e.g., $Task_{11}(s)$ calls $TaskType_1(s)$).

\paragraph{Reward functions:}
We define two reward functions.
On the task type level, the reward for proceeding with a subroutine from its current state $s$ with action $a$ is:
\begin{equation}
\begin{split}
	R_{t}(s,a) =
	\left\{
	\begin{matrix}
        - c_T(s) & \mathrm{if}~a~\mathrm{is}~leave\\
		r_T(s) & \mathrm{if}~a~\mathrm{is}~continue~,
	\end{matrix}
	\right.
\end{split}
\end{equation}
where $c_T(s)$ and $r_T(s)$ are the respective cost and reward functions of the task. 
This covers cases in which the agent gains a reward by pursuing a task ($r_T(s), a=continue$).
It also captures human sensitivity to future costs \cite{Altmann2002,McFarlane2002}, when deciding to terminate task execution ($-c_T(s), a=leave$). Finally, it models the effect of decreasing reward as well as increasing effort both increasing the probability of leaving a task \cite{Gutzwiller2019}.
On the task instance level, we penalize state changes to model reluctance to continue tasks that require excessive effort to recall relevant knowledge \cite{Altmann2007,Oulasvirta2006}.
The respective reward function is 
$R_{r}(s) = - c_T(z(s))$,
where $s$ is the state on the root-level, $z(s)$ maps $s$ to the state of its child's SMDP, 
and $c_T(s)$ is again the cost function of the task.



\paragraph{Hierarchical optimality:}
We model task interleaving as a \emph{hierarchically optimal} RL problem.
This captures the idea that rewards of the alternative tasks influence the decision to continue the attended task.
Therefore, we implement hierarchical task interleaving using the three-part value function decomposition proposed in \cite{Andre2002}.
The Bellman equation of a task type subroutine considers the expected reward outside of this routine and is defined as
\begin{align}
&Q^\pi_{t}(s,a) = \sum_{s'} P_{t}(s'|s,a)R_{t}(s,a) \\ \nonumber
&+ \sum_{SS(s',t)} F_{t}(t|s,a)P_{t}(s'|s,a)\gamma^t_{t} Q^\pi_{t}(s',\pi_{t}(s')) \\ 
&+ \sum_{EX(s',t)} F_{t}(t|s,a)P_{t}(s'|s,a)\gamma^t_{t} Q^\pi_{r}(p(s'),\pi_{r}(p(s')))~, \nonumber
\end{align}
where $s$, $a$, $s'$, $P_{t}$, $F_{t}$, $\pi_{t}$, and $\gamma_{t}$ are the respective functions or parameters of a task type level SMDP. $\pi_{r}$ is the optimal policy on the root level, and $p(s)$ maps from a state $s$ to the corresponding state in its parent's SMDP.
$EX(s',t)$ is a function that returns the subset of next states $s'$ and transition times $t$ that are exit states as defined by the environment of the subroutine. $SS(s',t)$ is a similar function that returns $s'$ and $t$ for all other states.
Note that on the lower level of the hierarchy the decision process is an SMDP rather than an MDP. Hence, we model 
varying progress speed (action times) per person and task type.
$Q^\pi_{r}$ is the Bellman equation on the root-level of our HRL-model and is defined as
\begin{align}
Q^\pi_{r}(s,a) & = \sum_{s',t} F_{r}(t|s,a)P_{r}(s'|s,a)[R_{r}(s) \\
& + Q^\pi_{t}(z(s),\pi_{type}(z(s)))+\gamma^t_{r} Q^\pi_{r}(s',\pi_{r}(s'))]~,\nonumber 
\end{align}
where $s$, $a$, $P_{r}$, $F_{r}$, $\pi_{r}$, and $\gamma_{r}$ are the respective functions or parameters of the root level SMDP. $z(s)$ is again the mapping function from root-level state to the state within its child's SMDP. State-action transition on the root level are rewarded according to the expected reward values of the subroutine $Q^\pi_{t}(z(s),\pi_{t}(z(s)))$ and penalized according to $R_{r}(s)$.
%
%
The discount factor $\gamma_{t}$ of $Q^\pi_{t}(s,a)$ can be used to model various degrees of executive control. A high executive control causes the agent to avoid switching to other tasks in anticipation of high future rewards and low executive control to switch for immediate rewards \cite{Wickens2008}. 
Using high respectively low values for $\gamma_{t}$ during training causes the agent to behave in a similar manner.
See \emph{Supplementary Material} for model details.

\subsection{Comparison with Flat RL}

\begin{figure}[tbh]
\centering
  \includegraphics[width=0.7\columnwidth]{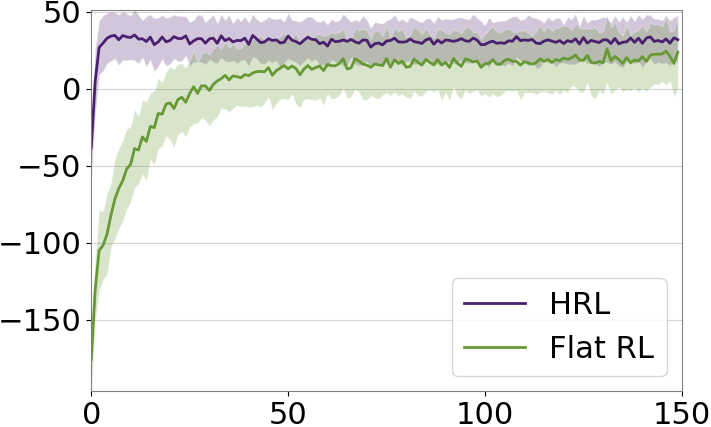}
  \caption{Learning curves of the flat RL- and our HRL-agent. Solid line denotes mean reward (y-axis) per episode (x-axis), shaded area represents standard deviation.}
  \label{fig:comparison_rl}
\end{figure}
To test the plausibility of hierarchicality, we further compared our model with a flat RL implementation of the task interleaving problem.
For both models, we learn 100 policies for a ten task, six instance problem and the same simulated user.
\figref{fig:comparison_rl} shows the learning curves of the two methods.
Both perform similarly in terms of attained reward per episode.
However, HRL converges faster than flat RL which is inline with prior work \cite{Dietterich1998,Andre2002,Ghavamzadeh2002}.
This is due to a significant decrease in the number of states (43-fold for this example).
This corroborates HRL as a cognitively more tractable model \cite{Botvinick2012,Frank2011}.

To find model-neutral evidence for the hierarchical assumption, we also analyzed the improvement of reward over time in the participant data of our experiment (see next section).
Here, reward increases quickly with experience over just three trials (1st trial: M 24.9 SD 12.4; 2nd: M 30.4 SD 11.5; 3rd: M 36.7 SD 11.1).
This can be attributed to participants ability, similarly as the HRL-agent, to generalize task instances to task types, which enables faster learning of reward estimates per task instance.
Flat RL does not generalize to task types, thus requires longer exposure.
\section{Experiments}
\label{sec:experiment}

We report results from (i) simulations and (ii) an interleaving study ($N=211$).
In all experiments, we trained the agent for 250 episodes\footnote{We consider an episode finished when all tasks in the task environment are completed.}, which was sufficient for saturation of reward. 
The HRL agent was trained using the discounted reward HO-MAXQ algorithm \cite{Andre2002}.

\begin{figure*}[t!]
\centering
\includegraphics[width=0.8 \linewidth]{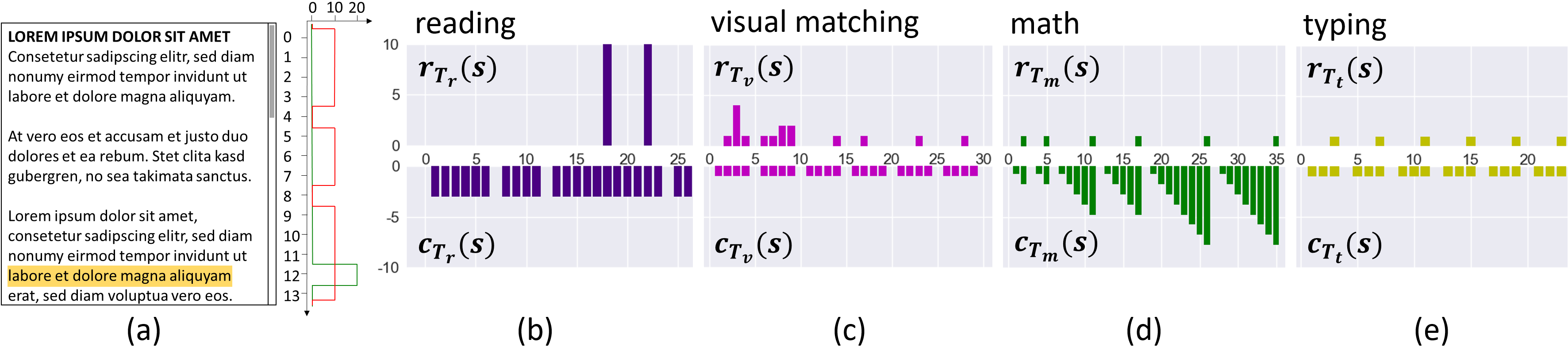}
\caption{Task models of the four tasks used in the experiment: (a) Example of how task state is assigned to visible state on display: passages of text in the reading task are assigned to the discrete states of its task model (column of numbers) over which reward (green) and cost function (red) are specified. The row highlighted yellow provides the answer to a comprehension query at the end. (b-e) show exemplary task models for (b) reading, (c) visual matching, (d) math, and (e) typing tasks.}
\label{fig:tasks}
\end{figure*}

\subsection{Simulations}

We report simulation results showing how the model adapts to changing cost/reward structures.
To this end, the two-task interleaving problem of \figref{fig:toyproblem} is considered.
The writing task $T_w$ awards a high reward when completed.
Switching away is costly, except upon completing a chapter.
The browsing task $T_b$, by contrast, offers a constant small reward and switch costs are low. 
In the following examples, the HRL agent was force to start with the writing task.

\paragraph{Cost and task boundaries:} In \figref{fig:toyproblem} c), the agent only switches to browsing \emph{after} reaching a subtask boundary in writing, accurately modelling sensitivity to costs of resumption \cite{Altmann2002,Gutzwiller2019,Iqbal2008}. 

\paragraph{Reward structure:} The HRL agent is sensitive to rewards \cite{Horrey2006,Iani2007,Norman1986,Wickens2008}, as shown by comparison of interleaving trajectories produced with different values of $\gamma$ in \figref{fig:toyproblem}. For example, when $\gamma = 0$, only immediate rewards are considered in RL, and the agent immediately switches to browsing.

\paragraph{Level of supervisory control:} The discount factor $\gamma$ approximates the level of executive control of individuals.
\figref{fig:toyproblem} d) illustrates the effect of high executive control: writing is performed uninterruptedly while inhibiting switches to tasks with higher immediate but lower long-term gains.

\subsection{Online interleaving study: Method}
Novel experimental data was collected to assess how well the model (i) generalizes to an unseen task environment and (ii) if it can account for individual differences. 
Participants practiced each task type separately prior to entering interleaving trials. 
Six task instances were made available on a browser view (see screen shots in \emph{Supp. Mat.}).
The reward structure of each task was explained, 
and users had to decide how to maximize points within a limited total time.

\paragraph{Participants:}  
218 participants completed the study. Ten were recruited from our institutions, 
and the rest from Amazon Mechanical Turk. 
Monetary fees were designed to meet and surpass the US minimum wage requirements.
A fee of 5 USD was awarded to all participants who completed the trial,
and an extra of 3 USD as a linear function of points attained in the interleaving trials.    
We excluded 7 participants who did not exhibit any task interleaving behavior.

\paragraph{Tasks:}
Four realistic task types were used.
(i) \emph{Reading} included snippets from an avalanche bulletin and two multiple choice questions to measure comprehension.
(ii) \emph{Typing} required transcribing six phrases displayed one at a time.
(iii) \emph{Math} required solving equations with addition and subtraction operators. 
Difficulty was increased by adding more terms.
(iv) \emph{Visual matching} consisted of six image lists,
and the participants had to click those showing airplanes. 
Difficulty was controlled by decreasing the proportion of images with airplanes. 
In-task rewards were designed to be realistic and clear. 
For instance, in Visual Matching, rewards were assigned per correctly identified airplanes.
Participants were told about the reward structures,
and feedback on attained rewards was provided. 

\paragraph{Procedure:}
After instructions, informed consent, and task type specific practice, the participants 
were asked to solve between two and five task interleaving trials.
Every trial contained six task instances, 
each sampled from a distribution of its general type. 
The instances were made available in a tabbed view.
A tab had to be clicked to resume the instance.
Trial durations were sampled from a random distribution unknown to the participant. 
The stated goal was to maximize total points linked to monetary rewards.
No task instance was presented more than once to a participant.
The average task completion time was 39 minutes.

\begin{figure*}[tbh]
\centering
\includegraphics[width=0.9 \linewidth]{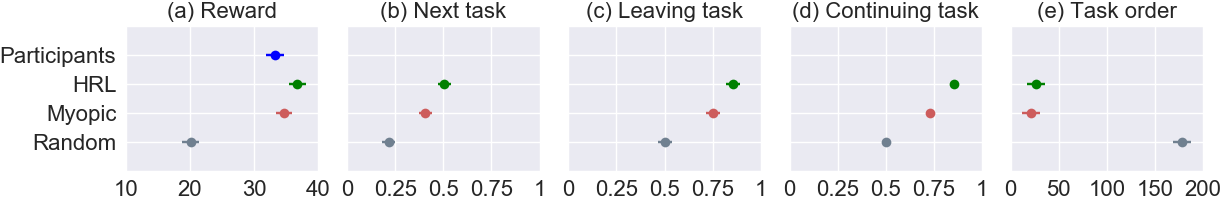}
\caption{
Means and 95\% confidence intervals for (a) attained rewards, (b) accuracy in predicting next task, (c) accuracy in predicting leaving of a task, (d) accuracy in predicting continuing of a task, (e) error in predicting order of tasks (lower is better).
}
\label{fig:results}
\end{figure*}

\subsection{Model fitting}

\paragraph{Task models:}
\figref{fig:tasks} shows task models that we used in the study.  
A mapping was created between what is shown on the display and task state $s$.
\figref{fig:tasks} a) shows an example: text paragraphs are mapped to the state of a reading model. 
Cost functions were designed to be constant or increasing linearly according to the designed difficulty in a task. 
For example, in the Math task, costs increased with the number of terms in an equation (see $c_{T_m}$ in \figref{fig:tasks} d).
Rewards were modeled based on the monetary reward, known to participants.
See \emph{Supp. Materials} for more details.

\paragraph{Empirical parameters:} The model predicts individual differences based on three types of idiosyncratic parameters:
(i) the discount factor $0.0 < \gamma_{t} < 1.0$,
(ii) a general switch cost $0.0 < c_{P} < 0.3$, 
(iii) and a task specific scaling coefficient $0.0 < s_{PT} < 1.0$.
Perceived switch cost, known to affect switching \cite{janssen2015strategic}, is modeled per task type as: $c_{PT}(s) = c_{P} + s_{PT}c_{T}(s)$. 

\paragraph{Inverse modeling method:} To fit these to an individual's data,
we used \emph{approximate Bayesian computation} (ABC) \cite{Kangasraasio2017,lintusaari2018}.
ABC is a sample-efficient and robust likelihood-free method for fitting simulator models to data. 
It yields a posterior distribution for the likelihood of parameter values given data.
An aggregate index of interleaving similarity is the to-be-minimized discrepancy function: 
\begin{align}
\label{eq:discrepancy}
d(S_{s}, A_{s}, \pi) &= - w \bigg[1 - \frac{1}{N_{s}} \sum_{s,a~\in~S_{s}, A_{s}} \mathds{1}(a, \pi(s))\bigg],\\ \nonumber
\begin{split}
\mathds{1}(a_p,a_a) &=
\left\{
\begin{matrix}
1 & \mathrm{if}~a_p=a_a\\
0 & \mathrm{else}~,
\end{matrix}
\right.
\end{split}
\end{align}
where $S_{s}$ is the set of states in which participants switched tasks, $A_{s}$ is the set of chosen actions (tasks) by the participant, $N_{s}$ is the number of task switches by the participant, $\pi$ is the hierarchical policy of the agent, and $w$ is a weight to scale discrepancy values (see \emph{Supp. Materials} for details). 

\paragraph{Fitting procedure:} 
We held out the \emph{last} trial of a participant for testing and used the preceding interleaving trials for parameter-fitting.
We run the above fitting method to this data for 60 iterations.
In each, we trained the HRL agent ten times using the same set of parameters in a task interleaving environment matching that of the participant in question.
%
For the Gaussian Process proxy model in ABC, we used a Matern-kernel parameterized for twice-differentiable functions.
%
On a commodity desktop machine (Intel Core i7 4GHz CPU), learning a policy took on average 10.3 sec (SD 4.0),
and fitting for full participant data took 103.8 min (SD 28.2).
The reported results come from the policy with lowest discrepancy to data obtained in 15 repetitions of this procedure with different weights (best: $w=100$).

\subsection{Baseline models}

We compare the HRL model against two baselines: 
\textbf{Random} chooses at each decision point of the SMDP at random one of the available actions;
\textbf{Myopic} chooses the task $T$ that provides the highest reward in its next state $s'$:
\begin{align}
&\pi^m = arg~max_T~r_T(s'_T) - \mathds{1}(T,T_{o})\bigg[c_T(s'_T) + c_T(s_{T_{o}})\bigg], \nonumber \\ 
\begin{split}
&\mathds{1}(T,T_{o}) =
\left\{
\begin{matrix}
0 & \mathrm{if}~T=T_{o}\\
1 & \mathrm{else}~,
\end{matrix}
\right.
\end{split}
\end{align}
where $s'_T$ is the next state of task $T$, $s_{T_{o}}$ is the current state of the ongoing task $T_{o}$, and $c_T$ is the respective task's cost function.
This baseline is a myopic version of the HRL model which only considers the next state.
To compare against a strong model, it decides based on the \emph{true} rewards and costs of the next states.
By contrast, HRL decides based on learned \emph{estimates}.
We did not compare against marginal rate of return \cite{Duggan2013} or information foraging models \cite{Payne2007} as in-task states can have zero reward.
Both models would switch task in this case, rendering them weaker baselines than Myopic. 
The multi-criteria model of \cite{Wickens2015} does not adapt to received task rewards and offers no implementation. 
Models of concurrent multitasking (i.e., \cite{Oberauer2011,Salvucci2008}) are not designed for sequential task interleaving.

\subsection{Results}

Predictions of HRL were made for the held-out trial and compared with human data. 
Analyzing base rates for continuing versus leaving a task of the behavioral sample revealed that task-continuation dominates events ($=0.95$). 
For this reason, we analyze the capability of models to predict if participants leave or continue a task separately. 

\paragraph{Reward:} HRL (M $36.7$, SD $9.51$) attained the highest reward, 
followed by Myopic (M $34.65$ SD $8.03$), Participants (M $33.18$, SD $11.92$) and Random
(M $20.08$, SD $8.52$). 

\paragraph{Choosing next task:} HRL showed highest accuracy in predicting the next task of a participant (M $0.51$, SD $0.27$), see \figref{fig:results} b). It was followed by Myopic (M $0.41$, SD $0.26$) and Random (M $0.22$, SD $0.21$). 
There was a significant effect of model ($H(2)=127.9,p<0.001$).%
\footnote{We use Kruskal-Wallis for significance-testing throughout.}
Post-hoc pairwise comparisons (Tukey's) indicated a significant difference between the models: all $p<0.001$.

\paragraph{Leaving a task:} HRL outperformed Myopic in predicting when a participant would leave a task (HRL, M $0.85$, SD $0.23$; Myopic, M $0.75$, SD $0.29$). Random was the worst (M $0.5$, SD $0.30$, see \figref{fig:results} c). 
There was a significant effect of model on accuracy ($H(2)=143.1,p<0.001$). 
Again, pairwise comparisons revealed a significant difference between all models ($p<0.001$). 

\paragraph{Continuing a task:} HRL (M $0.86$, SD $0.17$) was better than Myopic (M $0.73$, SD $0.22$) in predicting continuation in a task (see \figref{fig:results} d).
Random was the worst (M $0.50$, SD $0.07$). 
These differences were significant ($H(2)=327.0,p<0.001$). 
Pairwise comparisons indicated that significant differences hold between all conditions ($p<0.001$). 

\paragraph{Order of tasks:} 
HRL and Myopic were equally good in predicting the \emph{order} in which tasks are visited.
To this end, we defined \emph{task order error} as the sum of non-equal instances between produced orders of tasks.
A significant omnibus effect of model was found ($H(2)=346.2,p<0.001$). 
Myopic had a smaller error (M $20.60$, SD $29.43$) than HRL (M $26.21$, SD $29.74$). 
However, this difference was not statistically significant ($p=0.7$).
Random was the worst (M $177.90$, SD $113.19$). 
HRL and Myopic had a significantly smaller error than Random ($p<0.001$ for both).

\begin{figure}[tbh]
\centering
\includegraphics[width=1.0 \columnwidth]{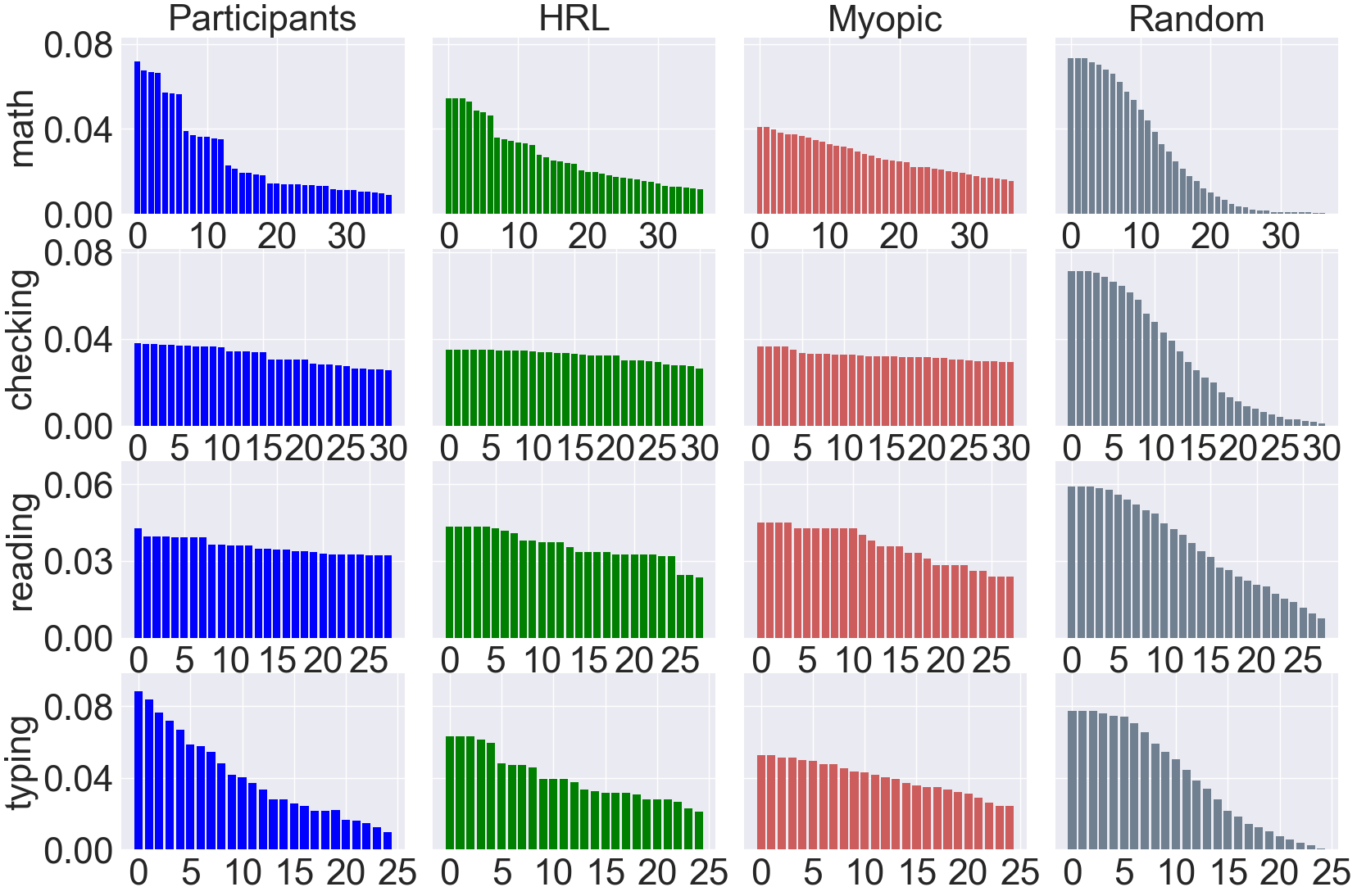}
\caption{
State visitations: HRL shows better match with state visitation patterns than Myopic and Random. y-axis shows fraction of states visited aggregated over all trials.
}
\label{fig:in-task-his}
\end{figure}

\paragraph{State visitations:} 
We computed histograms of state visitation frequencies per task type (\figref{fig:in-task-his}).
As visual inspection confirms, HRL had a superior histogram intersection ($0.93$) with Participants than Myopic ($0.88$) and Random ($0.81$).
The step-like patterns in the histograms of Participants were reproduced by HRL, illustrating that its policies switched at the same subtask boundaries as participants (e.g., see top-row in \figref{fig:in-task-his}).

\subsection{Parameter fitting}
\tabref{tab:fitting-metric} reports the mean fraction of reproduced actions per participant for each iteration of our model fitting procedure.
Fractions are computed using the normalized sum of reproduced actions of \equref{eq:discrepancy}.
Results on training trials improve with each iteration of the procedure and show that learned parameters generalize to the held-out test trials. 

\begin{table}[tbh]
\centering
	\begin{tabular}[c]{|l|l||l|l|}
	    \hline
		Trials&Iterations&Mean&Std. dev.\\
		\hline
		\hline
		training&random&0.51&0.21 \\
        &1&0.64&0.13 \\
        &10&0.69&0.13 \\
        &30&0.70&0.14 \\
		&60&0.73&0.13\\
		\hline
		test&random&0.54&0.21\\
		&fit&0.67&0.20\\
		\hline
	\end{tabular}
	\caption{Mean and standard deviation of fractions of reproduced actions of participants. Fractions are computed for fitted (test trial), random and inferred parameters after number of iterations (training trial).}
	\label{tab:fitting-metric}
\end{table}

The mean difference between the estimated parameters of two runs of the fitting procedure per participant are: $\gamma_{t}$ (discount factor) M $0.20$, SD $0.23$; $c_P$ (switch cost) M $0.09$, SD $0.09$; $s_{PR}$ (reading) M $0.32$, SD $0.28$; $s_{PV}$ (visual matching) M $0.36$, SD $0.28$; $s_{PM}$ (math) M $0.33$, SD $0.26$; 
$s_{PT}$ (typing) M $0.38$, SD $0.34$.
The somewhat low reliability of parameters can be explained by the fact that, in our task, participants can achieve a high reward by two means: high switch costs or high discount factor.
While our model parameters are theoretically justified, refining the model such that parameter values can reliably be associated with behavioral patterns is an interesting direction of future work.

\section{Discussion}

Understanding information processing in the brain requires computational models that are capable of performing realistic cognitive tasks by reference to neurobiologically plausible component mechanisms \cite{kriegeskorte2019}.
%
%
Computational models that synthesize task performance can expose interactions among cognitive components and thereby subject theories to critical testing against human behavior.
%
In this spirit, we have provided new evidence for hierarchical reinforcement learning (HRL) as a model of task interleaving.
The resemblance between simulated and empirical data is very encouraging. 
Comparison against the myopic baseline suggests that human interleaving is better described as optimal planning under uncertainty than by a myopic strategy.
We have shown that hierarchically optimal value decomposition is a tractable solution to the planning problem that the supervisory control system faces.
In particular, it
(i) can achieve a high level of control via experience,
(ii) adapts to complex and delayed rewards/costs, avoiding being dominated by immediate rewards, and
(iii) can generalize task type knowledge to instances not encountered previously.
Moreover, only a small number of empirical parameters was needed for characterizing individual differences. 

To support further research on the topic, we release our code and data as open source. 
One exciting remaining question is the relationship between HRL and heuristic strategies.
Our informal observations suggest that greedy behavior may emerge at the extreme when the discount factor approaches zero.
Another promising direction concerns the role of memory. 
With time-bound mechanisms in place, such as loss of activation,
we might see the time-based switching heuristic emerge.
We model recall effort to be the sole source of switch costs. 
For future work, it is interesting to extend cost functions to account for other factors, e.g., cognitive load.

\bibliographystyle{aaai}
\bibliography{references}

\end{document}